\documentclass{article}
\usepackage[a4paper]{geometry}
\usepackage[utf8]{inputenc}
\usepackage{amsfonts} 
\usepackage{amsmath}
\usepackage{systeme}
\usepackage{setspace}
\usepackage{cite}
\usepackage{setspace}
\usepackage[utf8]{inputenc}
\usepackage{fancyhdr}
\usepackage{graphicx}
\usepackage{mathtools}
\usepackage{pgfplots}
\usepackage{array}
\pgfplotsset{compat=1.15}
\usepackage{mathrsfs}

\usepackage{pgf,tikz}
\usetikzlibrary{arrows}
\usepackage{cases}
\usepackage{amssymb}
\usepackage{amsthm}
\usepackage{tabularx}
\usepackage{chessboard}
\usepackage{tikz-network}
\usepackage{tikz}

\usepackage{mathtools}
\newcommand{\tb}{\textbf}
\newcommand{\tl}{\textsl}

\newcommand{\nd}{\noindent}
\newcommand{\mc}{\mathcal}
\newcommand{\mb}{\mathbb}
\DeclareMathOperator*{\esssup}{ess\,sup}
\DeclareMathOperator*{\essinf}{ess\,inf}
\usetikzlibrary{shapes,backgrounds}
\usetikzlibrary{decorations.markings}
\usepackage{verbatim}
\makeatother
\usepackage[inline]{asymptote}
\usepackage{extarrows}
\usepackage{hyperref}
\hypersetup{
    colorlinks,
    citecolor=black,
    filecolor=black,
    linkcolor=black,
    urlcolor=black
}

\title{Continuous Classification Aggregation}
\author{Zijun Meng}
\date{}

\begin{document}
\linespread{1.5}
\setlength{\baselineskip}{18pt}
\maketitle

\begin{center}
    \textbf{Abstract}
\end{center}
We prove that any optimal, independent, and zero unanimous fuzzy classification aggregation function of a continuum of individual classifications of $m\ge 3$ objects into $2\le p\le m$ types must be a weighted arithmetic mean. We also provide a characterization for the case when $m=p=2$.

\tableofcontents

\newpage
\section{Introduction}
In parallel with Arrow's impossibility theorem (1951) \cite{a1951} on aggregating preferences, Kasher and Rubinstein (1997) proposed and axiomatically studied in \cite{k1997} what is now called the group identification problem. In their paper, they studied a setting where a finite group of people have to decide a subset of themselves, every individual have an opinion on the subset. They showed that any symmetric, monotonic, independent, consensus, and liberal method of aggregating the opinions of this group of people must be strongly liberal. Samet and Schmeidler (2003) provided in \cite{s2003} a classification for symmetric, monotonic and independent aggregation rules. Miller (2008) extended in \cite{m2008} the model appeared in Kasher-Rubinstein and Samet-Schmeidler and studied the situation when the group of people have to choose multiple subsets of themselves. Other studies that follow up on Kasher-Rubinstein (1997) include but not limited to Dimitrov-Sung (2005) \cite{d2005}, Huoy (2006) \cite{h2006}, and Huoy (2007) \cite{h2007}.

\vspace{0.3cm}

Maniquet and Mongin (2016) provided in \cite{m2016} another generalization of Kasher-Rubinstein and Samet-Schmeidler. They analyzed the problem when a group of people have to classify a group of objects into a group of categories, they showed that an independent and surjective aggregation of crisp classifications of a finite number of individuals satisfying an unanimity condition must be a dictatorship. For a comprehensive treatment on collective preference, we refer to Sen (1970) \cite{s1970}.

\vspace{0.3cm}

Every piece of the aforementioned studies are crisp in a sense that adscription is absolute, a person (or an object) either totally belongs to a category or totally does not belongs to a category, and no ambiguity is allowed. In order to incorporate some sort of vagueness, we need a concept of fuzzy preferences. This is dealt by Dutta-Panda-Pattanaik (1986) in \cite{d1986} and Dutta (1987) in \cite{d1987}. For more background on fuzzy mathematics, we refer to Bede (2013) \cite{b2013}, Syropoulos-Grammenos (2020) \cite{s2020} and Syropoulos (2025) \cite{s2025}, although these are a bit distant from our topic of discussion. A textbook that is closer to our favour would be Dubois-Prade (1980) \cite{d1980}.

\vspace{0.3cm}

The group identification problem in the fuzzy setting has aroused a wide interest among economists. Studies in this problem include but not limited to Barrett-Pattanaik-Salles(1985) \cite{b1985}, García-Lapresta-Llamazares (2000) \cite{g2000}, Ballester-García-Lapresta (2008) \cite{b2008}, Cho-Park (2018) \cite{c2018}, and Alcantud (2019) \cite{a2019}. In particular, Fioravanti (2025) extended Maniquet-Mongin (2016) in \cite{f2025} to a fuzzy setting. They proved that an optimal, independent and surjective aggregation of fuzzy classifications of a group of individuals satisfying a weaker unanimity condition must be a weighted arithmetic mean.

\vspace{0.3cm}

The work mentioned above are all dealing with a group of finite number of individuals. A natural question is whether the same is true for a large economy, where the set of individuals is uncountable, because an answer to this question would provide a powerful theoretical foundation. Also, a model with a continuum of individuals is more compatible with other continuous economic models, thus making up for the remoteness of the discrete models in social choice theory from a wide range of economic models.

\vspace{0.3cm}

The fuzzy setting in Fioravanti (2025) \cite{f2025} allowed us to model the set of individuals as an unit interval in a clean way. In this paper, we prove the same result in the large economy setting. Apart from the technical side, we apply the ideas in Acz´el-Ng-Wagner (1984) \cite{a1984}, Fioravanti (2025) \cite{f2025} and Wagner (1982) \cite{w1982}. Nonetheless, we feel the need to emphasize that the application of the mathematical tools (particularly the Riesz representation theorem) to this group identification problem is novel.

\vspace{0.3cm}

As for the organization of our paper, we introduce our setting in \tb{Section \ref{model}}. After an illustration in \tb{Section \ref{eg}}, we state our results in \tb{Section \ref{results}}. Proofs can be found in the appendix.

\section{Acknowledgements}
The author would like to thank
\begin{itemize}
    \item his teacher Dong Zhang for checking this piece of work and provide fixes;
    \item his colleague Cheuk Fung Lau for providing suggestions for polishing the writing of this paper.
\end{itemize}

\section{Model}\label{model}
\nd Throughout this paper, we use $\lambda$ to denote the Lebesgue measure.

\vspace{0.3cm}

\nd We use the model and assumptions of Fioravanti (2025) \cite{f2025}, but we change the set of individuals from a finite set to an uncountable set. More precisely, let $I=[0,1]$ be the set of individuals, $X=\{x_1, \dots, x_m\}$ be the set of $m$ objects. A \textsl{fuzzy classification} is a map $c\colon X\to I^p$, where $p\ge 2$ is the total number of types, $c\left(x\right)_t$ is interpreted as the proportion of the object $x$ classified into type $t$. We require \[\sum\limits_{j=1}^m c\left(x_j\right)_t\ge 1\] (this surjectivity condition requires $m\ge p$) and \[\sum\limits_{t=1}^p c\left(x_j\right)_t=1.\] Denote by $\mc C$ the set of fuzzy classifications and \[\mc C^I=\{c^I=(c_i)_{i\in I}\mid c_i\in\mathcal C\,\forall i\in I\}\] be the set of \textsl{fuzzy classification profiles}.

\vspace{0.3cm}

\nd A \tl{fuzzy classification aggregation function (FCAF)} is a map $\alpha\colon\mc C^I\to\mc C$. Let $\mu$ be a probability measure on $I$, then the map $\alpha_\mu\colon\mc C^I\to\mc C$ generated by $\mu$, defined by \[\alpha_\mu(c^I)(x)=\int_Ic_i(x)\mu(di),\] is easily verified to be an FCAF, we call it a \tl{weighted arithmetic mean}. For example, if $w\in C(I,\mathbb R_{\ge 0})$ is a weight function such that $\int_Iw(i)\lambda(di)=1$ with the corresponding Lebesgue-Stieljes measure $\mu=\lambda_w$, where $C(I,\mathbb R_{\ge 0})$ is the space of nonnegative continuous functionals on $I$, then $\alpha_\mu\colon\mc C^I\to\mc C$ given by \[\alpha_\mu(c^I)\left(x\right)=\int_Iw(i)c_i(x)\lambda(di)\] is a weighted arithmetic mean. Furthermore, we say that $\alpha_\lambda$ is an \tl{arithmetic mean}. Also, for any $i\in I$, we say that $\alpha_{\delta_i}$ is a \tl{degenerated weighted arithmetic mean}\footnote{Before choosing to model a weighted arithmetic mean using a probability measure, the author tried to use a weight function $w\in L^2I$. Although it works for our main result \tb{Theorem 1}, it does not work well for this concept and some results related to which.}, where $\delta_i$ is a probability measure assigning a point mass on $i$.

\vspace{0.3cm}

\nd\tb{Assumption 1} (Optimality)\tb{.} For any fuzzy classification $c^I\in\mathcal C^I$, for any type $1\le t\le p$, if \[\sum\limits_{j=1}^m c_i\left(x_j\right)_t=h\] for $\lambda$-a.e. $i\in I$, where $h\ge 1$ is a constant, then \[\sum\limits_{j=1}^m \alpha(c^I)\left(x_j\right)_t=h.\]

\vspace{0.3cm}

\nd\tb{Assumption 2} (Independence)\tb{.} For any fuzzy classifications $c^I,(c')^I\in\mathcal C^I$, for any object $x\in X$, if $c_i(x)=c'_i(x)$ for $\lambda$-a.e. $i\in I$, then $\alpha(c^I)(x)=\alpha((c')^I)(x)$.

\vspace{0.3cm}

\nd A stronger assumption than independence requires the FCAF to be insensitive to labels of objects:

\vspace{0.3cm}

\nd\tb{Assumption 2'} (Symmetry)\tb{.} For any fuzzy classifications $c^I,(c')^I\in\mathcal C^I$, for any objects $x, y\in X$, if $c_i(x)=c'_i(y)$ for $\lambda$-a.e. $i\in I$, then $\alpha(c^I)(x)=\alpha((c')^I)(y)$.

\vspace{0.3cm}

\nd\tb{Assumption 3} (Zero Unanimity)\tb{.} For any fuzzy classification $c^I\in\mathcal C^I$, for any type $1\le t\le p$ and object $x\in X$, if $c_i\left(x\right)_t=0$ for $\lambda$-a.e. $i\in I$, then $\alpha\left(c\right)\left(x\right)_t=0$.

\vspace{0.3cm}

\nd A stronger assumption than zero unanimity requires the FCAF to be unanimous at every point in $I$:

\vspace{0.3cm}

\nd\tb{Assumption 3'} (Unanimity)\tb{.} For any fuzzy classification $c^I\in\mathcal C^I$, for any type $1\le t\le p$ and object $x\in X$, for any $h\in I$, if $c_i\left(x\right)_t=h$ for $\lambda$-a.e. $i\in I$, then $\alpha(c)(x)_t=h$.

\vspace{0.3cm}

\nd An even stronger assumption requires the FCAF to be bounded by the essential upper and lower bounds:

\vspace{0.3cm}

\nd\tb{Assumption 3''} (Coherence)\tb{.} For any fuzzy classification $c^I\in\mathcal C^I$, for any type $1\le t\le p$ and object $x\in X$, we have \[\alpha(c)(x)_t\in\left[\essinf_{i\in I}c_i(x)_t,\esssup_{i\in I}c_i(x)_t\right].\]

\vspace{0.3cm}

\nd\tb{Assumption 4} (Non-dictatorship)\tb{.} There is no $i\in I$ such that $\alpha(c^I)=c_i$ for any $c^I\in\mathcal C^I$.

\vspace{0.3cm}

\nd A stronger assumption than non-dictatorship requires the FCAF to be insensitive to names of individuals:

\vspace{0.3cm}

\nd\tb{Assumption 4'} (Anonymity)\tb{.} For any fuzzy classifications $c^I, (c')^I\in\mathcal C^I$, for any $\lambda$-measurable subset $J\subseteq I$ such that $J+s\subseteq I$, and that $J$ and $J+s$ are disjoint, if $c_j(x)=c'_{j+s}(x)$ and $c_{j+s}(x)=c'_j(x)$ for any $j\in J$ and $x\in X$, then $\alpha(c^I)=\alpha((c')^I)$.

\vspace{0.3cm}

\nd The following proposition summarizing the relationships between the assumptions is immediate:

\vspace{0.3cm}

\nd\tb{Proposition 1.} The followings are true:
\begin{itemize}
    \item Symmetry $\implies$ independence;
    \item Coherence $\implies$ unanimity $\implies$ zero unanimity;
    \item Anonymity $\implies$ non-dictatorship.
\end{itemize}

\vspace{0.3cm}

\nd Next, we have the independence of the weakest versions of the four assumptions:

\vspace{0.3cm}

\nd\tb{Proposition 2.} For each of the assumptions among optimality, independence, zero unanimity and non-dictatorship, there exist an FCAF violating that assumption but satisfying the other three.

\section{An Illustration}\label{eg}
\nd\tb{Example 1.} Suppose $(m,p)=(6,3)$, and \begin{align*}c(x_1)(\cdot)&=(c(x_1)_1(\cdot),c(x_1)_2(\cdot),c(x_1)_3(\cdot))=\left(\frac23(\cdot),\frac13,\frac23(1-\cdot)\right),\\c(x_2)(\cdot)&=(c(x_2)_1(\cdot),c(x_2)_2(\cdot),c(x_2)_3(\cdot))=((1-\cdot)^3,3(\cdot)(1-\cdot),\cdot^3)),\\c(x_3)(\cdot)&=(c(x_3)_1(\cdot),c(x_3)_2(\cdot),c(x_3)_3(\cdot))=(\delta_0(\cdot),1-\delta_0(\cdot)-\delta_1(\cdot),\delta_1(\cdot)),\\c(x_4)(\cdot)&=(c(x_4)_1(\cdot),c(x_4)_2(\cdot),c(x_4)_3(\cdot))=(1,0,0),\\c(x_5)(\cdot)&=(c(x_5)_1(\cdot),c(x_5)_2(\cdot),c(x_5)_3(\cdot))=(0,1,0),\\c(x_6)(\cdot)&=(c(x_6)_1(\cdot),c(x_6)_2(\cdot),c(x_6)_3(\cdot))=(0,0,1).\\\end{align*}

\begin{center}
\begin{figure}[h]
\centering
\includegraphics[scale=10]{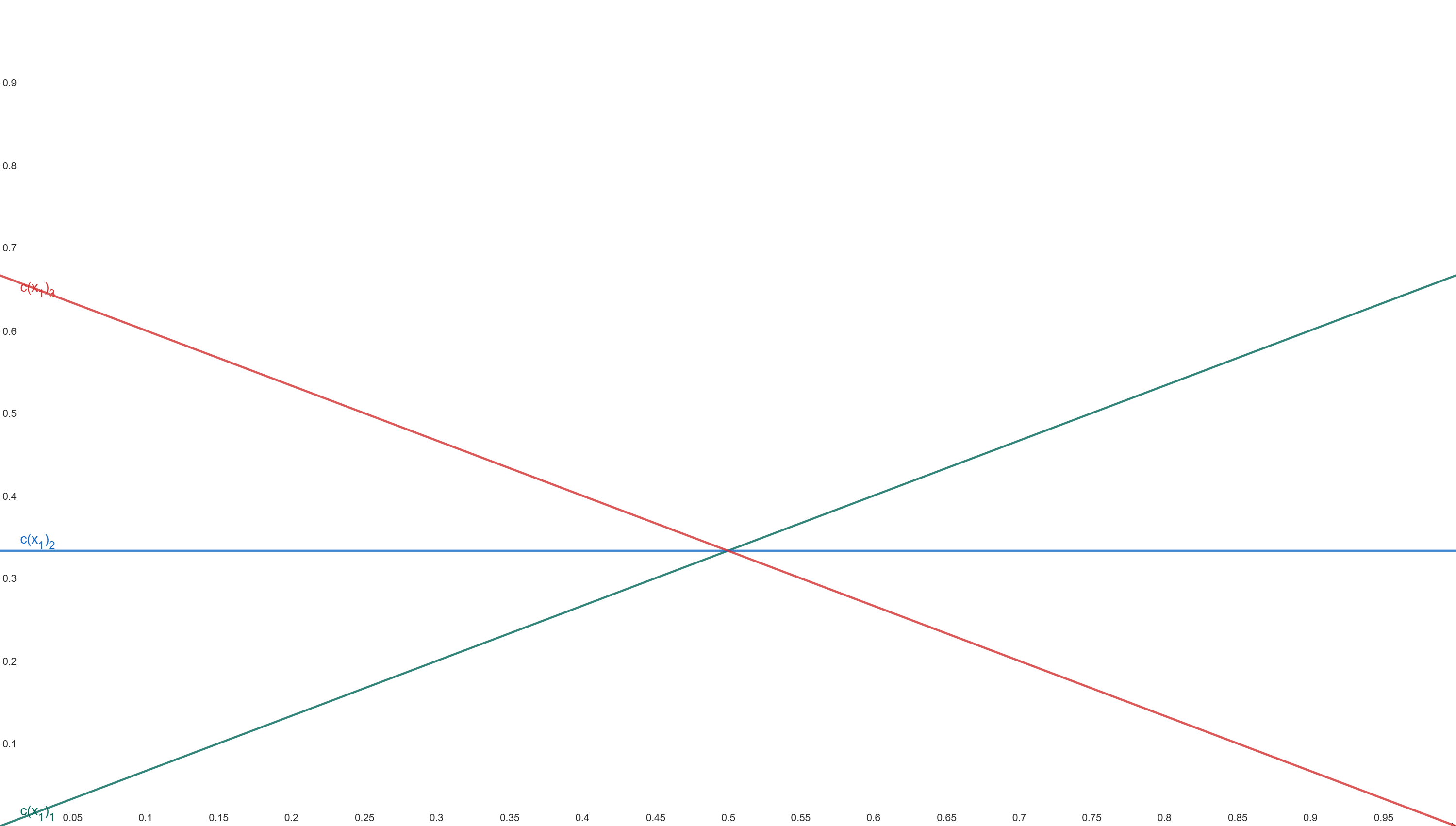}
\caption{$c(x_1)$.}
\end{figure}

\vspace{0.3cm}

\begin{figure}[h]
\centering
\includegraphics[scale=10]{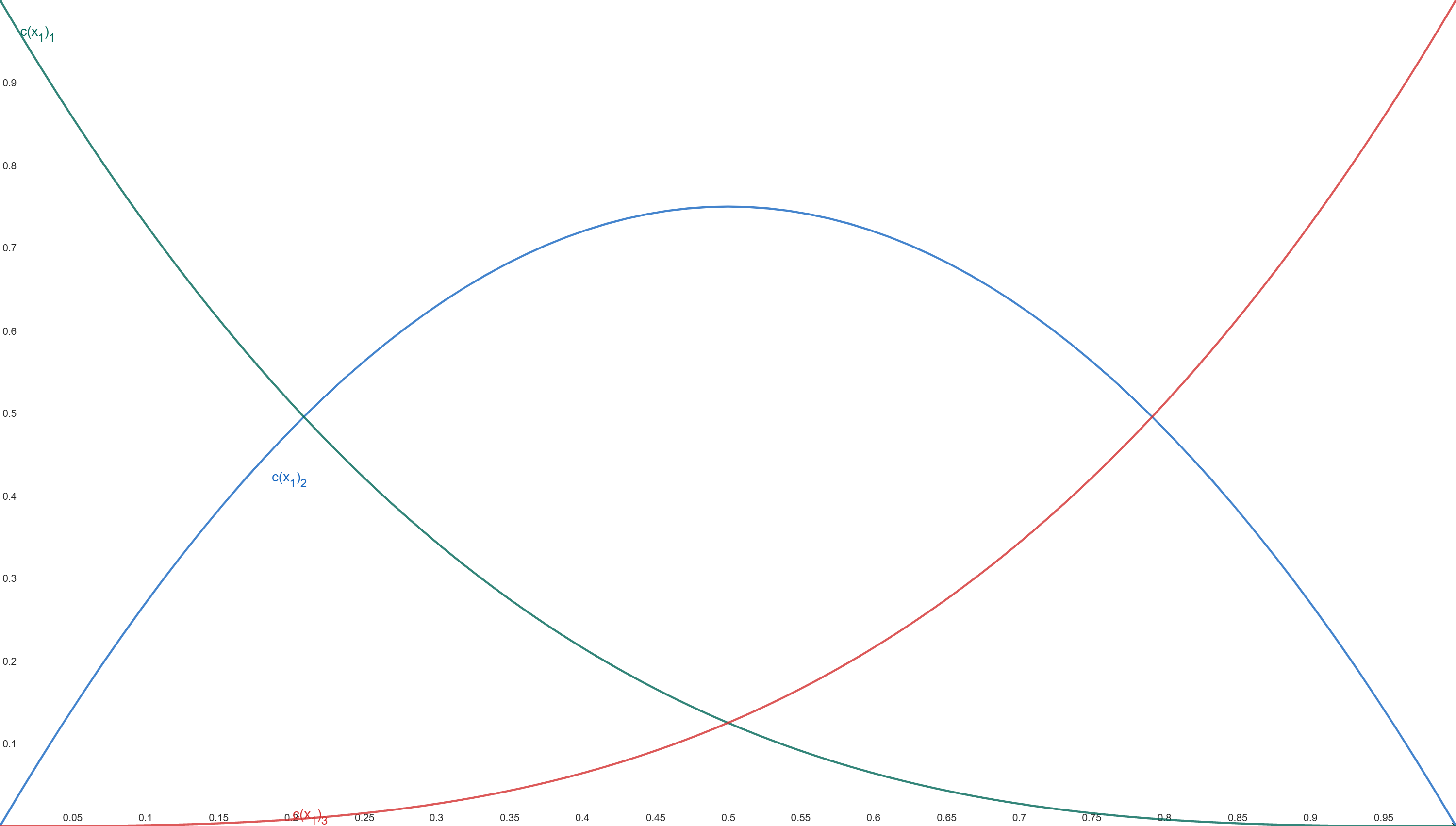}
\caption{$c(x_2)$.}
\end{figure}
\end{center}

\nd We can check that this is fuzzy because \[\sum\limits_{j=1}^6 c\left(x_j\right)_t\ge 1, \sum\limits_{t=1}^3 c\left(x_j\right)_t=1.\] Now if we use the weight arithmetic mean with the weight function $w(i)=3i^2$, we get the social classification \begin{align*}\alpha_w(c)(x_1)&=\left(\frac12,\frac13,\frac16\right),\\\alpha_w(c)(x_2)&=\left(\frac1{20},\frac9{20},\frac12\right),\\\alpha_w(c)(x_3)&=(0,1,0),\\\alpha_w(c)(x_4)&=(1,0,0),\\\alpha_w(c)(x_5)&=(0,1,0),\\\alpha_w(c)(x_6)&=(0,0,1).\\\end{align*}

\section{Results}\label{results}
To obtain our result, we use versions of the lemmas from \cite{a1984} and \cite{w1982} that are adapted to our setting. All proofs can be found in the appendix.

\vspace{0.3cm}

\nd\tb{Lemma 1.} If $\psi\colon [-\epsilon, \epsilon]^{[0,1]}=\{f\colon [0,1]\to[-\epsilon,\epsilon]\}\to\mb R$ is continuous with respect to the metric induced by the supremum norm, and satisfying the Cauchy equation \[ \psi\left(f+g\right)=\psi\left(f\right)+\psi\left(g\right)\,\forall f,g,f+g\in[-\epsilon,\epsilon]^{[0,1]},\] then there exists a continuous continuation $\overline\psi\colon C(I)\to\mb R$ of $\psi$ (i.e. such that $\overline\psi\vert_{[-\epsilon,\epsilon]^{[0,1]}}=\psi$) satisfying the Cauchy equation on $C(I)$: \[\overline\psi\left(f+g\right)=\overline\psi\left(f\right)+\overline\psi\left(g\right)\,\forall f,g\in C(I).\]

\vspace{0.3cm}

\nd\tb{Lemma 2.} Fix $m\ge 3$, a family of continuous maps $\{\alpha_j\colon I^I\to I\}_{j=1}^m$ satisfies the following two conditions \begin{align*}&\alpha_1(0)=\dots=\alpha_m(0)=0;\\&\forall f_1,\dots,f_m\in I^I, \sum\limits_{j=1}^mf_j=1\implies\sum\limits_{j=1}^m\alpha_j\left(f_j\right)=1\end{align*} iff there exists a probability measure $\mu$ on $I$ such that $\alpha_j(f)=\int_Ifd\mu$ for any $1\le j\le m$.

\vspace{0.3cm}

\nd\tb{Theorem 1.} Let $m\ge 3$. An FCAF $\alpha$ satisfies optimality, independence and zero unanimity iff it is a weighted arithmetic mean.

\vspace{0.3cm}

\nd With the previous theorem, we can immediately obtain the following corollaries:

\vspace{0.3cm}

\nd\tb{Corollary 1.} Let $m\ge 3$. An FCAF $\alpha$ satisfies optimality, independence, zero unanimity and anonymity iff it is a arithmetic mean.

\vspace{0.3cm}

\nd\tb{Corollary 2.} Let $m\ge 3$, $\alpha$ be an FCAF. Then the following statements are equivalent:
\begin{itemize}
    \item $\alpha$ is a non-degenerate weighted arithmetic mean;
    \item $\alpha$ satisfies optimality, independence, zero unanimity and non-dictatorship;
    \item $\alpha$ satisfies optimality, independence, unanimity and non-dictatorship;
    \item $\alpha$ satisfies optimality, independence, coherence and non-dictatorship;
    \item $\alpha$ satisfies optimality, symmetry, zero unanimity and non-dictatorship;
    \item $\alpha$ satisfies optimality, symmetry, unanimity and non-dictatorship;
    \item $\alpha$ satisfies optimality, symmetry, coherence and non-dictatorship.
\end{itemize}

\vspace{0.3cm}

\nd Our job would be complete if we can characterize the case when $m=2$. Note that in that case we must have $m=p=2$, and that any fuzzy classification \[c_i=(c^{x_1}_{i1},c^{x_1}_{i2},c^{x_2}_{i1},c^{x_2}_{i2})\in\mathcal C\] is fully parametrized by a single number (say $c^{x_1}_{i1}$) by the surjectivity assumption $\sum\limits_{j=1}^m c\left(x_j\right)_t\ge 1$, so any fuzzy classification when $m=p=2$ is automatically optimal. We begin with a lemma from \cite{w1982} adapted in our setting.

\vspace{0.3cm}

\nd\tb{Lemma 3.} A mapping $\alpha\colon([0,1]^I)^2\to[0,1]^2$ satisfies the following two conditions that for any $f_1,f_2\in [0,1]^I$, there is some $a_1, a_2\in[0,1]$ such that \begin{align*}&\alpha(f_1,f_2)=\alpha(f_2,f_1)\\&\alpha(f_1,0)=(a_1,0),\,\alpha(0,f_2)=(0,a_2)\end{align*} iff there exists a function $h\colon\left[-\frac12,\frac12\right]^I\to\left[-\frac12,\frac12\right]$ satisfying $h(\frac12)=\frac12$ and that $h(-f)=-h(f)$ for any $f\in\left[-\frac12,\frac12\right]^I$ such that for any $f_1,f_2\in[0,1]^I$, \[\alpha(f_1,f_2)=\left(h\left(f_1-\frac12\right)+\frac12,h\left(f_2-\frac12\right)+\frac12\right).\]

\vspace{0.3cm}

\nd\tb{Theorem 2.} Let $m=2$. An FCAF $\alpha$ satisfies symmetry and zero unanimity iff there is a function $h\colon\left[-\frac12,\frac12\right]^I\to\left[-\frac12,\frac12\right]$ satisfying $h(\frac12)=\frac12$ and $h(-f)=-h(f)$ for any $f\in\left[-\frac12,\frac12\right]^I$ such that \[\alpha(c^I)(x)=\left(h\left(c^I(x)_1-\frac12\right)+\frac12,h\left(c^I(x)_2-\frac12\right)+\frac12\right)\] for any $x\in X$.

\appendix
\section{Proofs}
This section is devoted to providing detailed proofs of all the theorems presented in this paper. Figure \ref{fig:logic} describes the logic flow between our results.

\tikzstyle{startstop} = [ellipse, rounded corners, minimum width=1cm, minimum height=1cm,text centered, draw=black, fill=white]
\tikzstyle{startstop2} = [ellipse, minimum width=1cm, minimum height=1cm,text centered, draw=black, fill=gray!10]
\tikzstyle{startstop3} = [rectangle, minimum width=1cm, minimum height=1cm,text centered, draw=black, fill=white]
\tikzstyle{startstop4} = [rectangle, minimum width=1cm, minimum height=1cm,text centered, draw=black, fill=gray!10]

\begin{figure}
    \centering
\begin{tikzpicture}[scale=1,node distance=3.9cm]
\node(L1)[startstop3]{Lemma 1};
\node(L2)[startstop3,right of=L1,yshift=0cm,xshift=0cm]{Lemma 2};
\node(T1)[startstop3,right of=L2,yshift=0cm,xshift=0cm]{Theorem 1};
\node(C1)[startstop3,right of=T1,yshift=0cm,xshift=0cm]{Corolloary 1};
\node(C2)[startstop3,above of=T1,yshift=-2cm,xshift=0cm]{Corolloary 2};
\node(L3)[startstop3,below of=L2,yshift=2cm,xshift=0cm]{Lemma 3};
\node(T2)[startstop3,right of=L3,yshift=0cm,xshift=0cm]{Theorem 2};
\draw[decoration={markings,mark=at position 1 with{\arrow[scale=2,>=stealth]{>}}},postaction={decorate}](L1) to node[anchor=south]{ } (L2);
\draw[decoration={markings,mark=at position 1 with{\arrow[scale=2,>=stealth]{>}}},postaction={decorate}](L2) to node[anchor=south]{ } (T1);
\draw[decoration={markings,mark=at position 1 with{\arrow[scale=2,>=stealth]{>}}},postaction={decorate}](T1) to node[anchor=south]{ } (C1);
\draw[decoration={markings,mark=at position 1 with{\arrow[scale=2,>=stealth]{>}}},postaction={decorate}](T1) to node[anchor=south]{ } (C2);
\draw[decoration={markings,mark=at position 1 with{\arrow[scale=2,>=stealth]{>}}},postaction={decorate}](L3) to node[anchor=south]{ } (T2);
\end{tikzpicture}
    \caption{Logic Flow}
    \label{fig:logic}
\end{figure}

\subsection{Proof of Proposition 2}
\begin{proof}
We provide the constructions one by one:
\begin{itemize}
    \item (Violating optimality) The FCAF $\alpha$ given by $\alpha(c^I)(x)=c_0(x)$ when $c_0(x)=e_k$ for some $1\le k\le p$ and \[\alpha(c^I)(x)=\left(\frac1p,\dots,\frac1p\right)\] otherwise, where \[e_k=\left(\underbrace{0,\dots, 0}_{k-1},1,0,\dots,0\right)\in\mb R^p.\]
    \item (Violating independence) Any FCAF $\alpha$ such that $\alpha(c^I)=c_0$ for any $c^I\in\mathcal A$ and $\alpha(c^I)=c_1$ for any $c^I\in\mathcal B$, where $\{\mathcal A,\mathcal B\}$ is a partition of $\mathcal C^I$.
    \item (Violating zero unanimity) The FCAF $\alpha$ given by $\alpha(c^I)(x_1)=c_0(x_2)$, $\alpha(c^I)(x_2)=c_0(x_1)$ and $\alpha(c^I)(x)=c_0(x)\,\forall x\in X\setminus\{x_1,x_2\}$ for any $c^I\in\mathcal C^I$. (Note that $\alpha$ passes independence and fails symmetry).
    \item (Violating non-dictatorship) The FCAF $\alpha$ given by $\alpha(c^I)=c_0$ for any $c^I\in\mathcal C^I$.
\end{itemize}
\end{proof}

\subsection{Proof of Lemma 1}
\begin{proof}
Note that for any $kf\in [-\epsilon,\epsilon]^{[0,1]}, k\in\mb Z$, we have \[\psi\left(kf\right)=k\psi\left(f\right),\] so, for any $g\in[-\epsilon,\epsilon]^{[0,1]}, k\in\mb Z$, we have \[\psi\left(g\right)=k\psi\left(\frac gk\right).\] Now, note that any $f\in C(I)$ is bounded, so there exists $k\in\mb Z$ such that $\frac fk\in[-\epsilon,\epsilon]^{[0,1]}$, define $\overline\psi\left(f\right)=k\psi\left(\frac fk\right)$. (From the above equality, this does not depend on the choice of $k$.) Substituting $k=1$ gives $\overline\psi\vert_{[-\epsilon,\epsilon]^{[0,1]}}=\psi$. Also, we have, for any $f,g\in C(I)$, \[\overline\psi\left(f+g\right)=k\psi\left(\frac{f+g}k\right)=k\left(\psi\left(\frac fk\right)+\psi\left(\frac gk\right)\right)=\overline\psi\left(f\right)+\overline\psi\left(g\right).\]
\end{proof}

\subsection{Proof of Lemma 2}
\begin{proof}
The if part is obvious. For the only if part, define \[\tilde I=\left[-\frac1m,1-\frac1m\right],\, \tilde f_j=f_j-\frac1m,\, \tilde\alpha_j(\tilde f_j)=\alpha_j\left(\tilde f_j+\frac1m\right)=\alpha_j\left(f_j\right).\] Now we have \[\left(\tilde f_j\in\tilde I^I\,\forall j, \sum\limits_{j=1}^m\tilde f_j=0\right)\implies\sum\limits_{j=1}^m\tilde\alpha_j(\tilde f_j)=1.\] In particular, \[\sum\limits_{j=1}^m\tilde\alpha_j\left(0\right)=1.\] Define $\psi_j(\tilde f)=\tilde\alpha_j(\tilde f)-\tilde\alpha_j\left(0\right)$. Then \[\left(\tilde f_j\in\tilde I^I\,\forall j,\sum\limits_{j=1}^m\tilde f_j=0\right)\implies\sum\limits_{j=1}^m\psi_j(\tilde f_j)=0,\psi_j\left(0\right)=0\,\forall j.\] In particular, \[\left(\tilde f_j\in[-\epsilon,\epsilon]^I\,\forall j,\sum\limits_{j=1}^m\tilde f_j=0\right)\implies\sum\limits_{j=1}^m\psi_j(\tilde f_j)=0,\psi_j\left(0\right)=0\,\forall j.\] For any $1\le j<k\le m$, $g\in[-\epsilon,\epsilon]^I$, substituting $\tilde f_j=g, \tilde f_k=-g$ and $\tilde f_\ell=0$ for all $j\ne\ell\ne k$, we get \[\psi_j\left(g\right)+\psi_k\left(-g\right)=0.\] Thus, for any pairwise distinct $j$, $k$, $\ell$, \[\psi_j\left(g\right)=-\psi_k\left(-g\right)=-\left(-\psi_\ell\left(-\left(-g\right)\right)\right)=\psi_\ell\left(g\right)\] Let $\psi=\psi_j\vert_{[-\epsilon,\epsilon]^I}\,\forall j$, then $\psi\left(g\right)+\psi\left(-g\right)=0$. Now, substituting \[\tilde f_1=f,\,\tilde f_2=g,\,\tilde f_3=-f-g,\,\tilde f_4=\dots=\tilde f_m=0,\] we know that $\psi$ satisfies the Cauchy functional equation for all $f,g,f+g\in[-\epsilon,\epsilon]^I$.

By \tb{Lemma 1}, we can uniquely extend $\psi$ to $\overline\psi\colon C(I,\tilde I)\to\mb R$ satisfying the Cauchy functional equation for all $f,g\in C(I)$, note that we still have $\overline\psi\ge 0$. Now, define \[\varphi_j(\tilde f)=\psi_j(\tilde f)-\overline\psi(\tilde f)\] for all $\tilde f\in\tilde I^I$, we show that $\varphi_j=0$ on $\tilde I^I$ for any $j$. Indeed, we know that $\varphi_j=0$ on $[-\epsilon, \epsilon]^I$. From the above, we get \[\left(\tilde f_j\in\tilde I^I\,\forall j,\sum\limits_{j=1}^m\tilde f_j=0\right)\implies\sum\limits_{j=1}^m\psi_j(\tilde f_j)=0.\] Hence, \[\sum\limits_{j=1}^m\varphi_j(\tilde f_j)=-\sum\limits_{j=1}^m\overline\psi(\tilde f_j)=-\overline\psi\left(\sum\limits_{j=1}^m\tilde f_j\right)=-\overline\psi\left(0\right)=0.\] For any $f\in\tilde I^I$, since $-f\in\left[-\frac{m-1}m,\frac1m\right]^I$, there exist $f_2,\dots, f_m\in\left[-\frac1m,1-\frac1m\right]^I=\tilde I^I$ such that \[f+\sum\limits_{j=2}^mf_j=0,\] and note that \[\frac1{m-1}\sum\limits_{j=2}^mf_j\in\tilde I^I.\] The above implication gives \[\varphi_1\left(f\right)+\varphi_2\left(\frac{-1}{m-1}f\right)+\dots+\varphi_m\left(\frac{-1}{m-1}f\right)=0.\] For any $\ell\ge 1$, replacing $f$ by $\left(\frac{-1}{m-1}\right)^\ell f$, we obtain \[\varphi_1\left(\left(\frac{-1}{m-1}\right)^\ell f\right)+\varphi_2\left(\left(\frac{-1}{m-1}\right)^{\ell+1}f\right)+\dots+\varphi_m\left(\left(\frac{-1}{m-1}\right)^{\ell+1}f\right)=0.\]

We know that for sufficiently large $\ell+1$, $\varphi_2, \dots, \varphi_m$ vanish, so the equality implies that \[\varphi_1\left(\left(\frac{-1}{m-1}\right)^\ell f\right)=0\] as well, so by symmetry \[\varphi_j\left(\left(\frac{-1}{m-1}\right)^\ell f\right)=0\] for any $j$. Recursively, we show that $\varphi_j\left(f\right)=0$ for any $j$, so $\psi_j=\overline\psi$ on $\tilde I^I$ for any $j$. Now note that $I$ is a locally compact Hausdorff space. Hence, due to the continuities of the $\alpha_j$'s, $\overline\psi$ is continuous as well. Since $\overline\psi$ satisfies the Cauchy functional equation, it follows that $\overline\psi\in(C(I,\tilde I))^*$. Recall that $\overline\psi\ge 0$, so we can apply a version of Riesz representation theorem (see, for example, theorem 2.14 of \cite{r1987}) and conclude that \[\overline\psi(f)=\int_Ifd\mu\,\forall f\in C(I,\tilde I)\] for some positive measure $\mu$ on $I$. It is easy to see that our conditions require $\mu$ to be a probability measure. Reverting, we obtain that for any $f\in C(I,I)$, \begin{align*}\alpha_j\left(f\right)&=\alpha_j\left(f\right)-\alpha_j\left(0\right)\\&=\tilde\alpha_j\left(f-\frac1m\right)-\tilde\alpha_j\left(-\frac1m\right)\\&=\psi_j\left(f-\frac1m\right)-\psi_j\left(-\frac1m\right)\\&=\overline\psi\left(f-\frac1m\right)-\overline\psi\left(-\frac1m\right)\\&=\int_I\left(f-\frac1m\right)d\mu-\int_I\left(-\frac1m\right)d\mu\\&=\int_Ifd\mu,\end{align*} concluding the proof.
\end{proof}

\subsection{Proof of Theorem 1}
\begin{proof}
The if part is obvious. For the only if part, notice that by independence we can decompose $\alpha$ into a set \[\{\alpha_{x_j}\colon\mathcal C^I\vert_{x_j}\to\mathcal C\vert_{x_j}\}_{j=1}^m\] of $m$ maps, each of which restricted on a single object $x_j\in X$. By optimality and zero unanimity, for each $1\le t\le p$, the family $\{\alpha_{x_j}\vert_t\}_{j=1}^m$ of maps satisfies the conditions of \tb{Lemma 2}, so applying the lemma, it follows that, there is a probability measure $\mu^t$ on $I$ such that \[\alpha_{x_j}\vert_t(c(x_j)_t)=\int_Ic_i(x_j)_t\mu^t(di).\] Now, it suffices to show that the measures $\mu^t$ are consistent between types $1\le t\le p$. Indeed, for any $1\le t\le p$, $\lambda$-measurable $J\subseteq I$, consider \[c_j^t\left(x_1\right)=\left(\underbrace{0,\dots, 0}_{t-1},1,0,\dots,0\right)\,\forall j\in J,\] and \[c_i^t\left(x_1\right)=\left(1,0,\dots,0\right)\,\forall i\in I\setminus J.\] Then, \begin{align*}\alpha (c^t)(x_1)&=\left(\int_{I\setminus J}c_i^t(x_1)_1\mu^1(di),\underbrace{0,\dots, 0}_{t-2},\int_Jc_j^t(x_1)_t\mu^t(dj),0,\dots,0\right)\\&=\left(\int_{I\setminus J}\mu^1(di),\underbrace{0,\dots, 0}_{t-2},\int_J\mu^t(dj),0,\dots,0\right),\end{align*} so \[1=\int_{I\setminus J}\mu^1(di)+\int_J\mu^t(dj).\] For any $1\le t'\le p$, comparing the last equation and the corresponding equation \[1=\int_{I\setminus J}\mu^1(di)+\int_J\mu^{t'}(dj),\] we obtain \[\int_J\mu^t(dj)=\int_J\mu^{t'}(dj).\] Since this equation is true for any $\lambda$-measurable set $J\subseteq I$, it follows that $\mu^t=\mu^{t'}$, establishing the consistency. 
\end{proof}

\subsection{Proof of Lemma 3}
\begin{proof}
The if part is obvious. For the only if part, from the symmetric and zero unanimous conditions imposed on $\alpha$, there exists a function $\tilde h\colon[0,1]^I\to[0,1]$ such that \[\alpha(f_1,f_2)=(\tilde h(f_1),\tilde h(f_2))\] and $h(0)=0$. Now, it is easy to see that $h(f)=\tilde h(f+\frac12)$ defined on $\left[-\frac12,\frac12\right]^I$ satisfies our requirements.
\end{proof}

\subsection{Proof of Theorem 2}
\begin{proof}
The if part is obvious. For the only if part, by zero unanimity and symmetry, for any $1\le t\le 2$, the map $\alpha\vert_t$ restricted on type $t$ satisfies the conditions of \tb{Lemma 3}. Applying \tb{Lemma 3}, there exists a function $h^t\colon\left[-\frac12,\frac12\right]\to\left[-\frac12,\frac12\right]$ satisfying $h(\frac12)=\frac12$ and $h(-f)=-h(f)$ for any $f\in\left[-\frac12,\frac12\right]$ such that for any $f_1,f_2\in[0,1]^I$, \[\alpha\vert_t(f_1,f_2)=\left(h^t\left(f_1-\frac12\right)+\frac12,h^t\left(f_2-\frac12\right)+\frac12\right).\] It is thus sufficient to show the consistency $h^1=h^2$. To see this, note that \begin{align*}h^1\left(c^I(x_1)_1-\frac12\right)+\frac12+h^2\left(c^I(x_1)_2-\frac12\right)+\frac12=1,\\h^1\left(c^I(x_1)_1-\frac12\right)+\frac12+h^1\left(c^I(x_2)_1-\frac12\right)+\frac12=1.\end{align*} It then follows from $c^I(x_1)_2=c^I(x_2)_1$ that $h^1=h^2$, as desired.
\end{proof}

\end{document}